\documentclass[11pt,a4paper]{article}
\usepackage[hyperref]{emnlp2020}
\usepackage{graphicx}
\usepackage{times}
\usepackage{latexsym}
\usepackage{linguex}
\usepackage{diagbox}
\usepackage{amsmath}

\usepackage{microtype}

\aclfinalcopy 

\title{Mention Extraction and Linking for SQL Query Generation}

\author{Jianqiang Ma\thanks{\; Equal contributions.} ,   
Zeyu Yan\footnotemark[1] , 
Shuai Pang, 
Yang Zhang, 
Jianping Shen\\
\{majianqiang554,yanzeyu751,pangshuai550,zhangyang147,shenjianping324\}@pingan.com.cn \\
AI Team, Ping An Life Insurance Company of China, Ltd\\
}

\date{}

\begin{document}
\maketitle
\begin{abstract}
On the WikiSQL benchmark, state-of-the-art text-to-SQL systems typically take a slot-filling approach by building several dedicated models for each type of slots. 
Such modularized systems are not only complex but also of limited capacity for capturing inter-dependencies among SQL clauses. 
To solve these problems, this paper proposes a novel extraction-linking approach, where  a unified extractor recognizes all types of slot mentions appearing in the question sentence  before a linker maps the recognized columns to the table schema to generate executable SQL queries. 
Trained with automatically generated annotations, the proposed method achieves the first place on the WikiSQL benchmark. 
\end{abstract}

\section{Introduction}
Text-to-SQL 
 systems generate SQL queries according to given natural language questions. Text-to-SQL technology is very useful as it can empower humans to naturally interact with relational databases, which serve as foundations for the digital world today. As a subarea of semantic parsing~\cite{Berant2013SemanticPO}, text-to-SQL is known to be difficult due to the flexibility in natural language. 

Recently, by the development of deep learning, significant advances have been made in text-to-SQL. 
On the WikiSQL \cite{Zhong2018Seq2SQLGS} benchmark for multi-domain, single table text-to-SQL, state-of-the-art systems~\cite{Hwang2019ACE,He2019XSQLRS} can predict more than 80\% of entire SQL queries correctly.  
Most of such systems take a sketch-based approach \cite{Xu2018SQLNetGS} that builds several specialized modules, each of which is dedicated to predicting a particular type of slots, such as the column in \texttt{SELECT}, or the filter value in \texttt{WHERE}.  
Such dedicated modules are complex and often fall short of capturing inter-dependencies among SQL sub-clauses, as each type of slots is modeled separately. 
To deal with these drawbacks, this paper formulates text-to-SQL as mention extraction and linking problems in a sequence labeling manner (Section \ref{sec:task}). In this new formulation, the key to synthesizing SQL is to extract the \textit{mentions} of SQL slots and the \textit{relations} between them.
Consider the question and its corresponding SQL query in example \ref{eg:1}, with the headers in the schema being \{\textsc{Lane}, \textsc{Name}, \textsc{Nationality},
\textsc{Split (50m)}, \textsc{Time}\}.
\ex. \label{eg:1}
\a. \textbf{Question}:  What is the total sum of 50m splits for Josefin Lillhage in lanes above 8?  
\b. \textbf{SQL}: \texttt{SELECT} \texttt{SUM}  (Split (50m)) \texttt{FROM} some$\_$table \texttt{WHERE}  Name = `Josefin Lillhage'  \texttt{AND} Lane $> 8$

\noindent We can see that many SQL elements, or \textit{slots}, such as \texttt{column} names of \textsc{Split (50m)} and \textsc{Lane}, \texttt{values} like ``Josefin Lillhage" and 8, as well as operators $>$ are mentioned with words similar in form and/or meaning. Moreover, the relations between the slot mentions,  such as $<$lanes, above, 8$>$ forming a filter condition, are represented by proximity in linear order or other linguistic cues.  
Thus, the recognition of the mentions and their relations would mostly reconstruct the intended SQL query from natural language 
question. 
\begin{figure*}[h!]
\centering
    \includegraphics[width=0.85\linewidth]{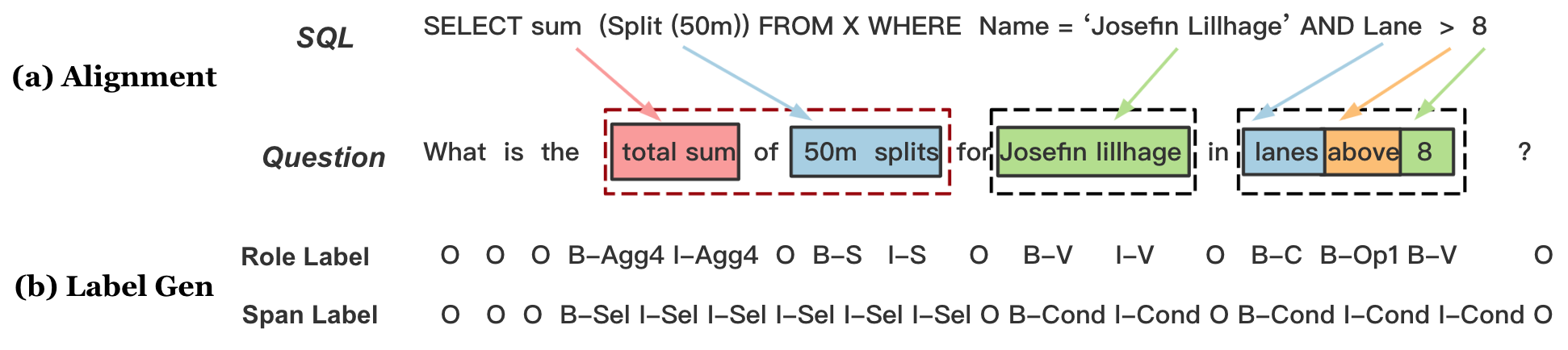}
    \caption{\textbf{Question sentence as mentions and relation of SQL slots.} (a) The mention of SQL slots (pink: the aggregation function, blue:columns, green: operators, yellow: values) and their relations (select relation and filter relation are shown in dashed rectangle).  (b) Representing mentions with role labels and relations with span labels.} 
    \label{fig:1}
\end{figure*}

To this end, we leverage one unified BERT-based \cite{Devlin2019BERTPO} extractor (Section~\ref{method:extraction}) to recognize the slot mentions as well as their relations, from the natural language questions. The output of the extractor can be deterministically translated into pseudo SQLs, before a BERT-based linker (Section \ref{task:linking}) maps the column mentions to the table headers to get executable SQL queries. 
A major challenge to the proposed method is the absence of manual annotation of mentions and relations. 
Thus we propose an automatic annotation method (Section \ref{method:annotation}) based on aligning tokens in a SQL with corresponding question. Also, preliminary results show that the prediction of aggregation function (AGG) restricts model performance, which induces us to put forward AGG prediction enhancement (AE) method inspired by \citet{brill1995transformation}. Trained with such annotations and applied AE method, the proposed method can already achieves the first place on the WikiSQL benchmark. 

The main contribution of this paper is the mention and relation extraction-based approach to text-to-SQL task. To the best of our knowledge, this is the \textit{first} work that formulates the task as sequence labeling-based extraction plus linking, which enjoys the advantage of structural simplicity and inter-dependency awareness.
In addition, we also propose an automatic method to generate annotations. 
Such annotations can be useful for developing novel methods for text-to-SQL, such as 
question decomposition-based approaches. 

\section{Method}\label{sec:task}
\subsection{Extractor}\label{method:extraction}
The extractor recognizes (1) \textit{slot mentions},  
including the SELECT column with aggregation function, WHERE columns with corresponding values and operators; and (2) \textit{slot relations}, namely associating each WHERE column with its operator and value. 
Most of the SQL slots are mentioned in the question, as shown in Figure \ref{fig:1}(a). 
As for the slot relations, 
 note that the column, value and operator that form a filter condition relation usually appear in adjacency in the question, such as \textit{in lanes above 8} in the example. Thus, the extraction of the relations is equivalent to the labeling of the corresponding \textit{text span}. 
As shown in Figure~\ref{fig:1}(b), the extraction of mentions and relations can be represented by tagging each token in the question by a set of BIO labels. 
Formally, the label $l \in \{T \times \{B, I\}, O\}$, where $\times$ denotes the Cartesian product of T, the set of functional labels, and the set of positional label of \{B, I\}, where 
  B and I means the beginning and the continuation of a particular annotation $t \in T$, respectively. The standing alone O label is assigned to tokens that are outside of any type of annotation of interest. 
For our task, we define two sets of labels: (a) the SQL \textit{role labels} representing the slot mentions; (b) the \textit{span labels} representing the slot relations, both of which 
 are shown in Table~\ref{table:mention}. 
With these defined label set, the recognition of both slot mentions and slot relations are formulated as \textit{sequence labeling}. 
\begin{table}[]
\centering
\begin{tabular}{l|l|llll}
\cline{1-3}
\textbf{Role Type} & \textbf{F-Labels}           & \textbf{Example} &  &  &  \\ \cline{1-3}
select column       & S                           & \textit{splits}: B-S      &  &  &  \\
where column        & C                           & \textit{lanes}: I-C       &  &  &  \\
value               & V                           & \textit{8}: B-V           &  &  &  \\
agg. function       & AGG$i$  & \textit{sum}: I-AGG4      &  &  &  \\
operator            & OP$i$   & \textit{above}:B-OP1      &  &  &  \\ \cline{1-3}
\cline{1-3}
\textbf{Span Type} & \textbf{F-Label} & \textbf{Example}                                                           &  &  &  \\ \cline{1-3}
SELECT span        & Sel              & 
\textit{sum}:I-Sel &  &  &  \\
FILTER span        & Cond             & \textit{lanes}:B-Cond 
&  &  &  \\ \cline{1-3}
\end{tabular}
\caption{\textbf{Labels for mention \textit{roles} \& relation \textit{spans}.}}
\label{table:mention}
\end{table}

\paragraph{Extractor Model}
The model first encodes the question text and the table headers. 
As pre-trained language models such as 
BERT achieve state-of-the-art performance on various NLP tasks including sequence labeling, we adopt BERT to get contextualized representations for both role and span labeling.  
Similar to state-of-the-art methods for text-to-SQL such as SQLova~\cite{Hwang2019ACE}, we concatenate the question text along with the table header as input for BERT,
in the form of $q_{1}, q_{2}, .., q_{L}, \texttt{[SEP]}, c_{1,1}, c_{1, 2}, ..., \texttt{[SEP]},$  $ c_{2,1},..., c_{M,1}...$,
 \ where $Q$ ($|Q| = L$) is the question while $C = c_1, .., c_M$ ($|C|=M$) are the table headers.  
Each header $c_i$ may have multiple tokens, thus the 2-d indexes of $c_{i,j}$ being used. 
Special \texttt{SEP} token is inserted between different headers $c_i$ as well as  between the question sentence $Q$ and the first header $c_1$. As the labeling is w.r.t. the question sentence, the conditional random filed (CRF) \cite{Lafferty2001ConditionalRF} layer only is applied to the question segment. The full model is described as in equation~(\ref{eq:model}), where BERT denotes the BERT model while CRF denotes a CRF layer.  
\begin{equation}\label{eq:model}
\begin{aligned}
Q^{\texttt{B}}; C^{\texttt{B}} &= \text{BERT} ([Q;C])\\
Q^{att} &= \text{Attention} (Q^{\texttt{B}}, C^{\texttt{B}}, C^{\texttt{B}}) + Q^{\texttt{B}} \\
L &= \text{CRF} (W^{L} Q^{att}) \\
\end{aligned}
\end{equation}
\noindent Before the BERT representations are fed to the CRF layer, they first go through an \textit{attention layer}~\cite{Bahdanau2014NeuralMT}, 
which encodes the question tokens with columns in the schema. The resulting representation is added to the original token representation in an element-wise manner. 
Finally, the resulting token representations are fed to the CRF layer, which yields the label sequence. 
As the two labeling tasks can benefit each other, we fine-tune BERT in a multi-task learning way. 

\subsection{Schema Linking as Matching} \label{task:linking} 
The column mentions in the question sentence often differ with the the canonical column names in the table schema in terms of string forms, as shown in  Figure~\ref{fig:1}, 
where \textsc{Split (50m)} is mentioned as \textit{50m splits} and \textsc{Name} is not mentioned at all. 
The latter case is \textit{implicit mention} of column, as only the value for the column, \textit{Josefin Lillhage}, appears in the question. Such case is challenging yet not uncommon. 
To convert mention and relation extraction results to SQL, we need a schema linking module to link explicit and implicit column mentions to its canonical column names in the table schema. 
Formally, we define the linker  
as a text matching model, i.e. estimating a function $f ([C_i;span;Q]) \rightarrow \{0,1\} $, where $C_i$ is a header in the table schema, $span$ is the either an extracted column mention (for linking explicit column mention) or an extracted value $v$ (for linking implicit column mention).
Special tokens of \texttt{[W]} and \texttt{[S]} are used to distinguish SELECT spans from FILTER spans. 
Again, BERT is used as the underlying model for its state-of-the-art performance on text matching. The matching procedure can be described as in equation~(\ref{eq:classify}). 

\begin{equation}\label{eq:classify}
\begin{aligned}
v^{\texttt{CLS}}_{i} &= \text{BERT} ([span;C_i])\\
P(i) &= \text{Sigmoid} (W v^{\texttt{CLS}}) 
\end{aligned}
\end{equation}

\subsection{AGG prediction enhancement} 
 Analysis of preliminary results suggests that aggregation function (AGG) prediction is a bottleneck for our system, which is partly attributed to the findings by  \citet{Hwang2019ACE} that AGG annotations in WikiSQL have up to 10\% of errors. 
 In such case,  as our extractor model has to take care of other types of slots, these extra constraints make it more challenging for our model to  
 fit flawed data, compared with a dedicated AGG classifier, as in most SOTA methods. 
 Another reason may be that no all the aggregation functions are grounded to particular tokens.  
 Given the characteristic of the data and the possible limitation of the information extraction-based model, we improve the AGG results over the original model, using only simple association signals in the training data. To this end, we adopt transformation-based learning algorithm~\cite{brill1995transformation} to update the AGG predictions based on association rules in the form of  ``change AGG from $x$ to $x'$, given certain word tuple occurrences." Such rules  are mined and ranked from the training data by the algorithm.

\subsection{Automatic Annotation via Alignment}\label{method:annotation}
A challenge for training the extractor is that 
benchmark datasets 
have no role or span annotations. 
Since manual annotations are costly, we resort to automatic ways. The idea is to annotate mentions by aligning the SQL slots in the query 
to tokens in the question.  Figure~\ref{fig:1} depicts such alignments 
with arrows and colors. 
Specifically, the proposed method is a two-step procedure. The first step is \textit{alignment}, 
which runs two pass of aligning. The first pass conducts exact and partial string match to recognize values and some of the columns, while the second pass aligns the remaining SQL slots, by training a statistical aligner with the training set of the data. For this purpose, we choose Berkeley aligner~\cite{liang2006alignment}, which 
works by estimating the co-occurrence of tokens in the parallel corpora, which are the question-SQL pairs in our case. 
As statistical aligner can occasionally yield null-alignment for a few tokens, 
we use another unsupervised word and semantic similarity-based algorithm~\cite{Perez2020UnsupervisedQD} to complement the missing alignments. 
The second step is \textit{label generation}, where the roles are generated according to aligned elements, while the span labels are assigned by considering minimal text span that covers all the elements in a SELECT/WHERE clause. 
\section{Experiment}
\textit{Dataset and Metric.} We use the largest human-annotated text-to-SQL dataset, WikiSQL \citet{Zhong2018Seq2SQLGS}, which consists of 80,654 pairs of questions and human-verified SQL queries. Tables appeared either in train or dev set will never appear in the test set. 
Two metrics 
 in \citet{Zhong2018Seq2SQLGS} are adopt for evaluating the SQL query synthesis accuracy:
(1) \textit{Logical Form Accuracy}, denoted as ${LF}$, 
where ${LF}$ = \# of SQL queries with correct logic form / total \# of SQL queries; and (2) \textit{Execution Accuracy}, denoted as ${EX}$,  
where ${EX}$ = \# of SQL queries with correct execution / total \# of SQL queries.
Execution guidance decoding (EG)
\citep{Wang2018ExecutionGuidedNP} is used, following previous work.

\textit{Implementation Details.} 
We use StanfordNLP \cite{Qi2018UniversalDP} for tokenization. The word embeddings are randomly initialized by BERT, and fine-tuned during the training. Adam is used  \cite{Kingma2014AdamAM} to optimize the model with default hyper-parameters. We choose uncased BERT-base pre-trained model with default settings due to resource limitations.  
The training procedures follows \citet{Hwang2019ACE}. Codes are implemented in Pytorch 1.3. 

\subsection{Results}
We compare our method with notable models that have reported results on WikiSQL task, including Seq2SQL\cite{Zhong2018Seq2SQLGS}, SQLNet\cite{Xu2018SQLNetGS}, TypeSQL\cite{Yu2018TypeSQLKT}, Coarse-to-Fine\cite{Dong2018CoarsetoFineDF}, SQLova\cite{Hwang2019ACE}, X-SQL\cite{He2019XSQLRS} and HydraNet~\cite{lyu2020hybrid} in Table~\ref{tab:main}. 
Without EG, our method with BERT-base outperforms most of existing methods, including SQLova with BERT-large and MT-DNN~\cite{Liu2019MultiTaskDN}-based X-SQL, 
and ranks right after HydraNet, which is based on RoBerTa~\cite{Liu2019RoBERTaAR} large. \citet{lyu2020hybrid} shows that RoBERTa large outperform BERT large in their setting and \citet{Liu2019MultiTaskDN} shows MT-DNN also outperforms BERT in many tasks.  Despite disadvantage in underlying pre-trained language model, our model achieves competitive results.  

\begin{table}[h!]
\hskip -0.0cm
\begin{tabular}{l|ll|ll}
\hline
\multicolumn{1}{c|}{Model} & \multicolumn{2}{c|}{Dev} & \multicolumn{2}{c}{Test} \\ \cline{2-5} 
\multicolumn{1}{c|}{}                       & LF      &  EX     &  LF      &  EX     \\ \hline
Seq2SQL                                     & 49.5        & 60.8       & 48.3        & 59.4       \\
SQLNet                                      & 63.2        & 69.8       & 61.3        & 68.0       \\
TypeSQL                                     & 68.0        & 74.5       & 66.7        & 73.5       \\
Coarse-to-Fine                              & 72.5        & 79.0       & 71.7        & 78.5       \\
SQLova                                & 81.6        & 87.2       & 80.7        & 86.2       \\
X-SQL                                       & 83.8        & 89.5       & 83.3        & 88.7       \\ 
HydraNet                                & 83.6  & \textbf{89.1}   & 83.8  & \textbf{89.2}    \\
this work - AE                                & 81.1        & 86.5       & 81.1        & 86.5       \\ 
this work                               & \textbf{84.6}        & 88.7       & \textbf{84.6}        & 88.8       \\ \hline
SQLova+EG                           & 84.2        & 90.2       & 83.6        & 89.6       \\
X-SQL+EG                               & 86.2        & 92.3       & 86.0        & 91.8       \\ 
HydraNet+EG                           & 86.6   & 92.4   & 86.5   & 92.2    \\
this work - AE$_{EG}$                             & 85.8        & 91.6       & 85.6       & 91.2   \\
this work$_{EG}$                            & \textbf{87.9}        & \textbf{92.6}       & \textbf{87.8}       & \textbf{92.5}     
\\ \hline
\end{tabular}
\caption{\textbf{Accuracy of previous and this work}.} 
\label{tab:main}
\end{table}
 For the results with the EG in Table~\ref{tab:main}, our method outperforms all the existing methods, including SQLova, X-SQL and HydraNet, leading to  
new state-of-the-art in the SQL accuracies in terms of both logic form and execution.
 Table~\ref{tab:aux} shows the slot type-wise results, where our method achieves new state-of-the-art results on the W$_{col}$, W$_{val}$ and W$_{op}$ accuracies.  
Since the operators and values are directly derived from the extractor, such results are evidence for the effectiveness of our extraction-based approach.
 Before applying AGG enhancement (\textit{AE}), the bottleneck of our method is on AGG prediction. We close such gap with \textit{AE} using only word co-occurrence features. The improved AGG accuracy also leads to the new state-of-the-art for the overall SQL results. Error analysis shows that our sequence-labeling-based model performs passably on some questions with nested structure. Consider the question ``When does the train [arriving at [Bourne] at 11.45] departure?", where the span for one condition ( \textit{Going to}=)  ``Bourne” is nested in the span for the other condition  \textit{arriving at} = 11:45. Such nested structure raises challenges to sequence labeling, similar to the situation encountered in nested NER. 

\begin{table}[h!]
\hskip -0.1cm
\centering
\tabcolsep=0.1cm
\begin{tabular}{l|llllll}
\hline
Model           & S$_{col}$     & S$_{agg}$     & W$_{no.}$     & W$_{col}$     & W$_{op}$      & W$_{val}$     \\ \hline
SQLova          & 96.8 & 90.6 & 98.5 & 94.3 & 97.3 & 95.4 \\
X-SQL           & 97.2 & 91.1 & \textbf{98.6} & 95.4 & 97.6 & 96.6 \\ 
HydraNet  &  \textbf{97.6} & 91.4 & 98.4 & 95.3 & 97.4 & 96.1 \\
ours-AE   & \textbf{97.6} & 90.7 & 98.3 & \textbf{96.4} & \textbf{98.7} & \textbf{96.8}    \\
ours  & \textbf{97.6}  &  \textbf{94.7}  & 98.3  &  \textbf{96.4} &  \textbf{98.7} &  \textbf{96.8}   \\ \hline
SQLova$_{EG}$       & 96.5 & 90.4 & 97.0 & 95.5 & 95.8 & 95.9 \\
X-SQL$_{EG}$        & 97.2 & 91.1 & \textbf{98.6} & 97.2 & 97.5 & 97.9 \\ 
HydraNet$_{EG}$  & \textbf{97.6}  &  91.4 & 98.4 & 97.2 &  97.5 & 97.6 \\ \hline
ours-AE$_{EG}$  &  \textbf{97.6}  & 90.7 & 98.3 & \textbf{97.9} & \textbf{98.5} & \textbf{98.3}  \\ 
ours$_{EG}$ & \textbf{97.6} & \textbf{94.7} & 98.3 & \textbf{97.9} &  \textbf{98.5} &  \textbf{98.3}  \\
\hline
\end{tabular}
\caption{\textbf{Test accuracy for each slot type.}}
\label{tab:aux}
\end{table}

\textit{Estimating Annotation Quality}. The quality of automatic annotation can be estimated in an \textit{oracle extractor} setting, where  
the automatically annotated labels, instead of the extractor prediction, are fed to the linker.  In this setting, the logic form and execution accuracy on the dev set reaches $92.8\%$ and  $94.2\%$, respectively,  which are the ceiling for our approach. Note that such ceiling 
is above the human-level accuracy reported in \citet{Hwang2019ACE},  
suggesting that the quality of the automatic annotation is reasonably good. 

\section{Related Work}
Semantic parsing~\cite{Berant2013SemanticPO} is to map natural language utterances to machine-interpretable representations, such as logic forms \cite{Dong2016LanguageTL}, program codes \cite{Yin2017ASN}, and SQL queries \cite{Zhong2018Seq2SQLGS}.
Text-to-SQL is a sub-area of semantic parsing, which is widely studied in recent years.
Earlier work \cite{Dong2016LanguageTL,Krishnamurthy2017NeuralSP,Zhong2018Seq2SQLGS,Sun2018SemanticPW,Wang2018ExecutionGuidedNP} follow a neural sequence-to-sequence paradigm \cite{Sutskever2014SequenceTS} with attention mechanism \cite{Bahdanau2014NeuralMT}. Pointer networks~\cite{Vinyals2015PointerN} are also commonly adopted. 
These sequence-to-sequence approaches often suffer the  ``ordering issue" since they are designed to fit an ordered sequence, while the conditions in WHERE-clause are unordered in nature.  

SQLNet~\cite{Xu2018SQLNetGS} introduces sketch-based method, which decomposes the SQL synthesis into several independent classification sub-tasks, including select-aggregation/column and where-number/column/operator/value. Except where-value, which is usually predicted by a pointer network, all the other sub-tasks use their own dedicated classifiers to make predictions. 
These sketch-based models raise challenges in training, deployment and maintenance.
Moreover, each sub-module 
solves its own classification problem, without considering the dependencies with SQL elements modeled by other sub-modules. 
Recent advances \cite{Yu2018TypeSQLKT,Dong2018CoarsetoFineDF,Hwang2019ACE,He2019XSQLRS} follow this approach and achieve comparative results on WikiSQL, 
 mostly by using pre-trained language models as the encoder.  

While our sequence labeling method is also based on  pre-trained language model, it differs from state-of-the-art methods in that it explicitly extracts mentions from the questions and can benefit from inter-dependency modeling between extracted mentions.  
The mentions for values, operators and corresponding columns often appear in proximity in the question, thus the sequence labeling model can better capture their dependencies and benefits the recognition for all of them, as experiment results suggest. 
Furthermore, our extractor-linker architecture is also much simpler than sketch-based methods.

Recent trend~\cite{Krishnamurthy2017NeuralSP,Guo2019TowardsCT,wang-etal-2020-rat,choi2020ryansql} in academia starts to shift to multi-table and complex queries setting of text-to-SQL, as in the Spider task~\cite{Yu2018SpiderAL}. State-of-the art methods on Spider typically fall into two categories: \textit{grammar-based approach}~\cite{Guo2019TowardsCT,wang-etal-2020-rat}, and \textit{sketch-based approach}, such as RYANSQL~\cite{choi2020ryansql} and RECPARSER~\cite{Zeng2020RECPARSERAR}. The latter ones have slot prediction modules similar to SQLNet for the WikiSQL, while recursion modules are introduced to handle the generation of complex SQL sketches, a characteristic in Spider but absent in WikiSQL. At a high level, our method is along the same line of  SQLNet-RYANSQL, yet differs with them, as our method extracts slots in a unified way rather than using dedicated modules to predict each slot type. 
We can extend our method to the Spider task by following existing sketch construction methods as in RYANSQL, while replacing their slot classification modules with our extractor-linker methods. 


\section{Conclusion and Future Work}
Thanks to the simple, unified model for mention and relation extraction and its capacity for capturing inter mention dependencies, the proposed method proves to be a promising approach to text-to-SQL task. 
Equipped with automatic-generated labels and AGG enhancement method, our model achieves state-of-the-art results on the WikiSQL benchmark. 
Since the current automatic-generated annotations are still noisy, it is useful to further improve the automatic annotation procedure. 
We also plan to extend our approach to cope with multi-table text-to-SQL task Spider.  

\section*{Acknowledgements}
We thank Jun Xu, Muhua Zhu, Wanxiang Che and Longxu Dou as well as all the anonymous reviewers for their invaluable comments and suggestions.


\bibliography{emnlp2020}
\bibliographystyle{acl_natbib}
\end{document}